\documentclass{article} 
\usepackage[preprint]{aistats2026}

\usepackage{amsmath,amsfonts,bm}

\def\eqref#1{equation~\ref{#1}}

\def\1{\bm{1}}

\DeclareMathAlphabet{\mathsfit}{\encodingdefault}{\sfdefault}{m}{sl}
\SetMathAlphabet{\mathsfit}{bold}{\encodingdefault}{\sfdefault}{bx}{n}

\usepackage{graphicx}
\usepackage{multirow}
\usepackage{makecell}
\usepackage{hyperref}
\usepackage{url}
\usepackage{booktabs}   
\usepackage{siunitx} 
\usepackage{graphicx}
\usepackage{multirow}
\usepackage{wrapfig}
\usepackage{amsthm}
\theoremstyle{plain}
\newtheorem{theorem}{Theorem}
\newtheorem{lemma}{Lemma}
\newtheorem{proposition}{Proposition}
\usepackage{tabularx}
\usepackage[table]{xcolor}

\newcolumntype{Y}{>{\centering\arraybackslash}m{0.1\linewidth}}
\usepackage{xr}
\title{GeoRecon: Graph-Level Representation Learning for 3D Molecules  via Reconstruction-Based Pretraining}

\begin{document}

\twocolumn[

\aistatstitle{GeoRecon: Graph-Level Representation Learning for 3D Molecules via Reconstruction-Based Pretraining}

\aistatsauthor{ Shaoheng Yan$^{1,2}$\And Zian Li$^{1,3}$\And Muhan Zhang$^{1,*}$ }

\aistatsaddress{$^1$Institute for Artificial Intelligence, Peking University\\ $^2$Yuanpei College, Peking University\\ $^3$School of Intelligence Science and Technology, Peking University } ]


\begin{abstract}
The pretraining–finetuning paradigm has powered major advances in domains such as natural language processing and computer vision, with representative examples including masked language modeling and next-token prediction. In molecular representation learning, however, pretraining tasks remain largely restricted to \emph{node-level} denoising, which effectively captures local atomic environments but is often insufficient for encoding the global molecular structure critical to graph-level property prediction tasks such as energy estimation and molecular regression. To address this gap, we introduce GeoRecon, a \emph{graph-level} pretraining framework that shifts the focus from individual atoms to the molecule as an integrated whole. GeoRecon formulates a graph-level reconstruction task: during pretraining, the model is trained to produce an informative graph representation that guides geometry reconstruction while inducing smoother and more transferable latent spaces. This encourages the learning of coherent, global structural features beyond isolated atomic details. Without relying on external supervision, GeoRecon achieves generally improves over backbones baselines on multiple molecular benchmarks including QM9, MD17, MD22, and 3BPA, demonstrating the effectiveness of graph-level reconstruction for holistic and geometry-aware molecular embeddings.
\end{abstract}
\section{Introduction}
With the pretraining–finetuning paradigm extending from computer vision and natural language processing ~\citep{bao2021beit,lu2019vilbert} to an increasingly broad range of domains, designing effective pretraining tasks that enable models to achieve competitive downstream performance with less data and faster convergence has become a key component in training pipeline design. In molecular representation learning, pretraining strategies based on denoising or masking have been widely validated for their effectiveness~\citep{Zaidi2023,zhou2023uni,feng2023fractional}. In these tasks, molecular structures are perturbed or masked at the \emph{atomic (\textbf{node}) level}, and the model is trained to recover the original data.

However, a substantial portion of downstream tasks in molecular representation learning require the model to possess a strong understanding at the \emph{whole-molecule (\textbf{graph}) level}~\citep{li2020learn}. Tasks such as energy or interatomic force prediction demand not only that the learned representation manifold faithfully capture node-level properties, but also that the model encode emergent graph-level information into the molecular representation. Unfortunately, conventional node-level pretraining tasks, by their very design, often lack explicit alignment with graph-level objectives, thereby limiting the potential gains that can be realized during the pretraining stage, motivating a shift from purely node-level perturbation to graph-level supervision.

We conducted a preliminary experiment by perturbing the atomic coordinates of a molecule with small displacements and using a spectral norm method described in \autoref{app:b} to estimate the Lipschitz constant of the encoder in the public Coord~\citep{Zaidi2023} pretraining checkpoint, a standard node-denoising-based baseline. This constant measures the sensitivity of the encoder’s output representation to coordinate changes (larger values indicate higher sensitivity).
\sisetup{
    scientific-notation = true,
    exponent-product = \times,
    round-mode = figures,
    round-precision = 4
}
\begin{table}[ht]
\centering
\caption{Local non-rigid Lipschitz constant \(L(x)\) for GeoRecon and the Coord baseline under two parameter settings. 
\texttt{steps} is the number of iterations in the power method used to estimate the dominant singular value.}\vspace{0.5em}
\label{Lipshitz}
\resizebox{\linewidth}{!}{
    \begin{tabular}{cc S[table-format=2.3e1] S[table-format=2.3e1] S[table-format=2.3e1]}
    \toprule
    Model & \texttt{step} & {5} & {15} & {25} \\
    \midrule
    \multirow{2}{*}{\textbf{GeoRecon}} 
    & \emph{median} & 2.969e1 & 3.071e1 & 3.075e1 \\
    & \emph{p95}    & 5.087e1 & 5.090e1 & 5.090e1 \\
    \midrule
    \multirow{2}{*}{\textbf{Coord}}
    & \emph{median} & 2.537e4 & 2.539e4 & 2.539e4 \\
    & \emph{p95}    & 4.629e4 & 4.638e4 & 4.451e4 \\
    \bottomrule
    \end{tabular}
    }
\end{table}

As shown in the second row of \autoref{Lipshitz}, Coord exhibits values exceeding $10^4$, indicating extreme sensitivity to minute perturbations. This instability underscores the limitations of node-level pretraining, as smooth representation manifolds are well known to facilitate downstream performance~\citep{lee2025enhancing,guo2024smooth,zhang2025repcali,krishnan2020lipschitz}. We provide a brief proof in \autoref{sec:smooth_finetune}.

To address such limitations, prior work has introduced explicit supervision via graph-level attributes~\citep{hu2019strategies} or frequent functional motifs~\citep{rong2020self}, or adopted contrastive learning~\citep{liu2021pre}. However, these methods are often over-specialized by hand-crafted supervision, limiting their general applicability, or are primarily designed for 2D or mixed 2D\&3D settings, lacking the inherent support for fully 3D molecular pretraining offered by denoising-based frameworks \citep{Zaidi2023}.

To further bridge the gap between node and graph-level pretraining, we propose \textbf{Geo}metric \textbf{Recon}struction, a graph-level 3D self-supervised pretraining task. Given a 3D molecular structure, the model is trained to \emph{produce an informative, orientation-invariant graph-level representation} that can guide the reconstruction of the molecule’s full 3D geometry from heavily noised coordinates. This forces the encoder to capture \emph{rich, molecule-level} structural information sufficient to recover heavily perturbed geometries.


GeoRecon possesses several key properties that make it distinct from prior graph-level pretraining approaches and broadly applicable across molecular modeling scenarios:
\begin{itemize}
    \item \textbf{3D-only input}: Relies \emph{solely} on atomic coordinates, without requiring 2D bond graphs or structural annotations, avoiding dependence on noisy or ambiguous 2D labels;
    \item \textbf{Fully self-supervised}: Learns directly from raw coordinates without any external labels;
    \item \textbf{Label-free and broadly applicable}: Does not require additional annotations (e.g., spectral data), enabling application to a wide range of molecular datasets;
    \item \textbf{Architecture-agnostic pretraining objective}: Can be integrated into different SE(3)-equivariant backbones without altering downstream fine-tuning protocols.
\end{itemize}
To demonstrate the effectiveness of GeoRecon, we fine-tune the pre-trained model on a range of graph-level molecular property prediction benchmarks, including QM9~\citep{ramakrishnan2014quantum}, MD17~\citep{chmiela2017machine}, and the challenging MD22~\citep{chmiela2023accurate} dataset. Our method \emph{consistently outperforms} our  backbone on MD17 and MD22, and achieves improvements on most metrics of QM9.

We further assess the robustness and generality of GeoRecon through two parallel evaluations: (i) ablations on training set size, encoder depth, pooling strategy, decoder depth, and reconstruction noise scale, and (ii) an out-of-distribution evaluation on 3BPA comparing GeoRecon with its TorchMD-Net backbone. Taken together, these results indicate that incorporating explicit graph-level structure in pretraining enhances both accuracy and robustness, with particularly strong benefits for challenging molecular modeling tasks that depend heavily on graph-level information.

\section{Related Works}
\textbf{Geometric Graph Neural Networks}\quad
Early methods incorporated 3D geometry into GNNs via continuous-filter convolutions and directional message passing. SchNet~\citep{schutt2018schnet} operates directly on Cartesian coordinates to predict energy and forces. DimeNet~\citep{gasteiger2020directional} and DimeNet++~\citep{gasteiger2020fast} capture angular information, improving performance and efficiency. GemNet~\citep{gasteiger2021gemnet} further enhances accuracy using multi-hop and higher-order geometric features. Meanwhile, EGNN~\citep{Satorras2021} and SE(3)-Transformer~\citep{fuchs2020se} encode Euclidean symmetries through equivariance, matching tensor-based models with lower complexity.

\textbf{Implicit Use of 3D Information}\quad
Some prior works implicitly leverage 3D structural information by converting 1D SMILES strings or 2D molecular graphs into 3D conformations using cheminformatics tools such as RDKit~\citep{liu2021pre,zhu2022unified,stark20223d}. These derived conformers are then used during training to enhance model capacity, achieving strong results on benchmarks like PCQM4Mv2~\citep{hussain2024triplet}, which are originally designed for 2D graph modalities.

\textbf{Self‑Supervised Denoising Pretraining}\quad
Self‑supervised denoising has emerged as a principled means to leverage unlabeled molecular conformations.
Coord pretraining \citep{Zaidi2023} demonstrates that training a GNN to remove Gaussian perturbations from equilibrium structures is mathematically equivalent to learning the underlying molecular force field, markedly improving downstream QM9 performance under limited labels. Frad \citep{feng2023fractional} furthers this concept by jointly perturbing dihedral angles and Cartesian coordinates, better emulating realistic conformational variations. SliDe \citep{ni2023sliced} augments denoising tasks with classical force‑field priors (bond lengths, angles, torsions), grounding the learned representation in established physical interactions. More recently, MolSpectra \citep{Wang2025} integrates multimodal energy‑level spectra via contrastive reconstruction, enriching the model’s access to discrete quantum information without external labels.

\textbf{Multi-Task and Multi-Fidelity Learning}\quad
To address data scarcity in quantum chemistry, one strategy is to exploit shared structure across tasks or fidelity levels. In \emph{multi-task learning}, models predict several quantum properties simultaneously (e.g., 12 QM9 targets), encouraging a shared embedding space and reducing overfitting. In \emph{multi-fidelity learning}, one leverages the hierarchy of computational costs: accurate methods like CCSD(T) are expensive, while cheaper ones like DFT scale to larger datasets. Recent work \citep{ren2023force} combines abundant DFT data with limited CC samples to approach CC-level accuracy at far lower cost. Together, these strategies mitigate scarcity by balancing accuracy and efficiency.

\textbf{Graph-Level Pretraining in Molecular Representation Learning}\quad
Early GNN pretraining mainly targeted node-level tasks such as masked atom prediction~\citep{hu2019strategies,hou2022graphmae,xia2023mole}, but recent efforts increasingly adopt graph-level objectives to capture molecule-wide signals. GROVER~\citep{rong2020self} introduces motif prediction to identify recurring substructures, and MGSSL~\citep{zhang2021motif} extends this by generating graphs motif-by-motif, injecting global inductive biases. GraphCL~\citep{you2020graph} and GCC~\citep{qiu2020gcc} apply graph-level contrastive learning with augmented graph views to enforce whole-graph consistency.

Other approaches leverage cross-modal or multi-scale alignment: GraphMVP~\citep{liu2021pre} aligns 2D graphs with 3D conformers via contrastive learning, while Uni-Mol2~\citep{ji2024uni} uses a two-track transformer jointly encoding atomic, graph, and spatial features. Though effective, these methods often require extra modalities, augmentation pipelines, or complex designs. In contrast, our method employs a single graph-level reconstruction objective—conditioned on pooled global embeddings—built directly atop coordinate denoising, achieving strong downstream performance without auxiliary data or heavy overhead.
\section{Preliminary}

Denoising has emerged as a principled and empirically effective pretraining objective in 3D molecular representation learning~\citep{Zaidi2023, feng2023fractional, zhou2023uni}. Formally, given a slightly perturbed molecule conformation $\widetilde{\bm{r}} = \bm{r} + \bm\epsilon  \in \mathbb{R}^{n\times 3}$ where $n$ denotes the molecule's node number, $\bm{r} \in \mathbb{R}^{n\times 3}$ the molecule equilibrium structure and $\epsilon$ the Gaussion noise, the model is asked to predict the noise from the perturbed conformation $\mathrm{GNN}(\widetilde{\bm{r}}) \approx \bm\epsilon$.
\citet{Zaidi2023} formally established the theoretical equivalence between coordinate denoising and force field prediction by leveraging the connection to score matching.
\begin{theorem}[Equivalence between denoising and force field prediction~\citep{Zaidi2023}]
\label{theo-1}
    \[\begin{aligned}
        \mathbb{E}_{q_\sigma(\widetilde{\bm{r}}, \bm{r})}&
    \left[
    \left\|
    \mathrm{GNN}_\theta(\widetilde{\bm{r}}) -
    \nabla_{\widetilde{\bm{r}}} \log q_\sigma(\widetilde{\bm{r}} \mid \bm{r})
    \right\|^2
    \right]
    \\&=
    \mathbb{E}_{q_\sigma(\widetilde{\bm{r}}, \bm{r})}
    \left[
    \left\|
    \mathrm{GNN}_\theta(\widetilde{\bm{r}}) -
    \frac{\bm{r} - \widetilde{\bm{r}}}{\sigma^2}
    \right\|^2
    \right].
    \end{aligned}
    \]
 \( \bm{r} \) denotes the equilibrium molecular structure, \( \widetilde{\bm{r}} \) is its perturbed version obtained by Gaussian corruption, and \( \theta \) represents the parameters of the denoising network. \( q_\sigma(\widetilde{\bm{r}} \mid \bm{r}) \) is defined as a Gaussian centered at \( \bm{r} \), serving as a tractable approximation to the Boltzmann distribution \( p_{\text{physics}}(\widetilde{\bm{r}}) \propto \exp(-E(\widetilde{\bm{r}})) \). Its score function corresponds to the standardized noise vector \( (\bm{r} - \widetilde{\bm{r}})/\sigma^2 \). More broadly, \( p_{\text{physics}} \) can be approximated by a Gaussian mixture \( q_\sigma (\widetilde{\bm{r}}) \sim \frac{1}{n} \sum_i \mathcal{N}(\widetilde{\bm{r}}; \bm{r}_i, \sigma^2 I) \), whose score defines a force field over noisy conformations.
    \end{theorem}
    
This theoretical result provides a rigorous foundation for coordinate denoising as a physically grounded pretraining objective, and is empirically validated by the Coord model.

Building upon coordinate denoising, \citet{feng2023fractional} introduced additional dihedral angle noise $\delta\psi \sim \mathcal{N}(0, \sigma^2 I_m)$ along rotatable bonds. 
While \citet{feng2023fractional} advance coordinate denoising by incorporating molecular flexibility through dihedral perturbations, their framework still focuses on node-level supervision. The model predicts local coordinate noise for each atom, but without explicit mechanisms to capture global structural context, which limits its effectiveness on graph-level tasks.
\section{Our Method}
\begin{figure}[h]
    \centering
    \vspace{-1em}
    \includegraphics[width=0.45\textwidth]{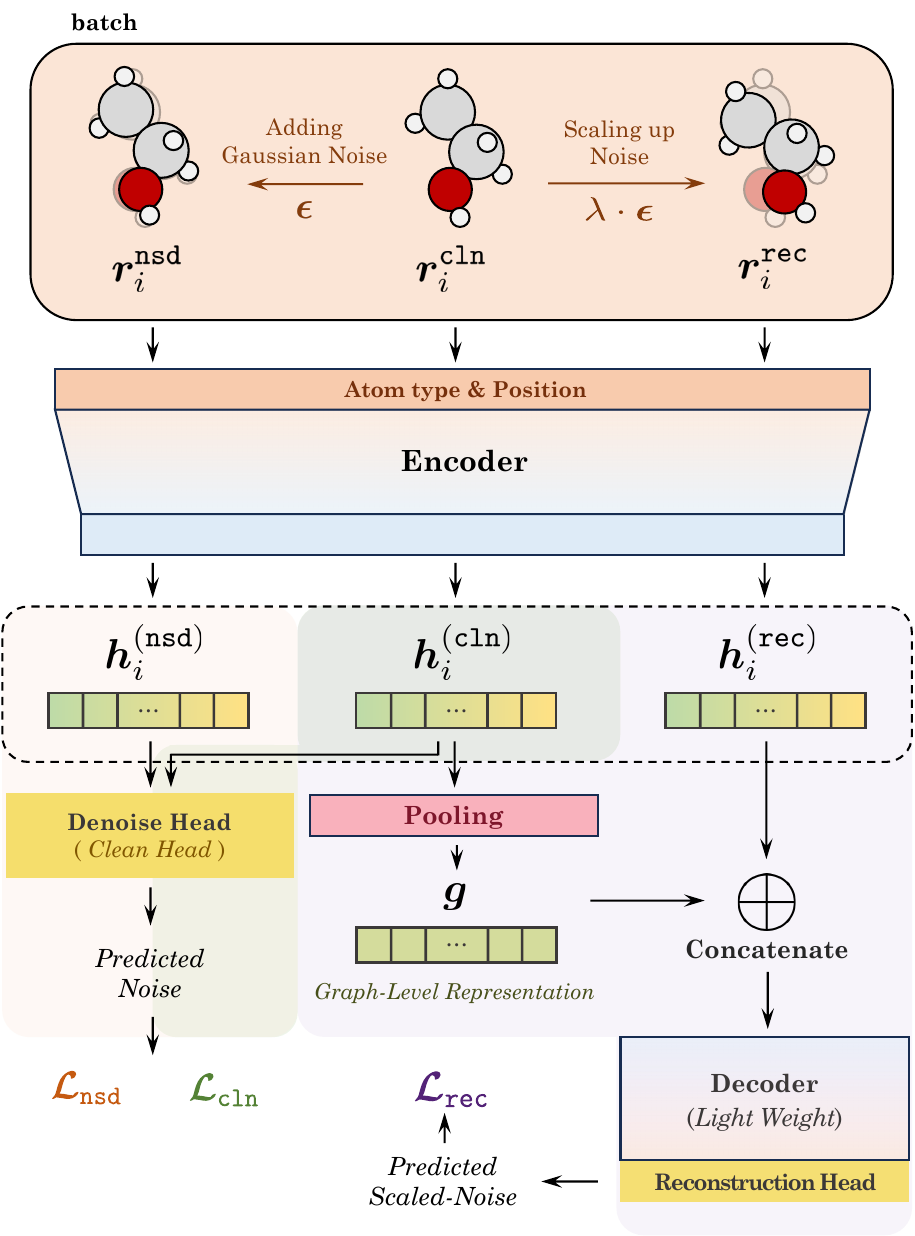}
   \caption{Overview of the GeoRecon framework. Given a molecular structure with atom types and 3D coordinates, the model encodes it using SE(3)-equivariant attention. Besides the standard node-level denoising objective, GeoRecon feeds a pooled graph-level representation concatenated with node embeddings derived from noisy coordinates into a lightweight decoder to reconstruct the scaled noise. The pretrained encoder is then finetuned for downstream molecular property prediction tasks.}
    \label{fig:model-wrap}
\end{figure}
Building on the self-supervised, physically grounded coordinate-denoising objective \citep{Zaidi2023,feng2023fractional}, we exploit its dual benefit as both a force-field surrogate and a fine-grained geometric bias. Yet, when used in isolation, this purely node-level signal cannot enforce the global structural coherence required by graph-level property prediction. {This gap motivates the need for a complementary mechanism that explicitly encourages the encoder to capture molecule-wide dependencies rather than only local perturbations.}

We therefore introduce \textbf{GeoRecon}, a reconstruction-based pretraining framework designed to capture global molecular structure from noisy inputs. A naive way is to directly decode coordinates from pooled graph-level embeddings. However, this method often loses spatial granularity. Furthermore, predicting equivariant coordinates from invariant graph embeddings means overfitting to specific orientations. Therefore, we formulate reconstruction as a denoising process too over entire molecular geometries. Specifically, we inject scaled Gaussian noise into the 3D coordinates and train a lightweight decoder, \emph{conditioned on graph-level representations}, to predict the scaled noise.

More precisely, GeoRecon extends standard coordinate denoising by increasing the task difficulty and conditioning the decoder on graph-level representations. Unlike approaches that carefully adjust noise to match specific physical semantics \citep{feng2023fractional,liu2025denoisevae}, we deliberately add \emph{orientation-invariant} noise into the coordinates and reduce the layer-num of the decoder. This places greater reliance on the global embedding, compelling the encoder to capture long-range dependencies and coherent molecule-level geometry. As a result, GeoRecon learns more transferable representations that bridge local interactions with global structural awareness.

\subsection{Notations}
\label{notations}
To support multi-task supervision, we construct three coordinate variants per molecule. (i) \textsc{Pos\_clean}: the original coordinates, serving as the ground truth for the denoising task and simultaneously as the input for the reconstruction encoder task introduced in the following section, denoted by \( \bm{r}_i^{\texttt{cln}} \in \mathbb{R}^{ 3} \), where \( i \) indexes the atom in the molecule; (ii) \textsc{Pos\_noised}: the perturbed atomic coordinates used as input for the standard node-level denoising task, denoted by \( \bm{r}_i^{\texttt{nsd}} \in \mathbb{R}^{ 3} \); (iii) \textsc{Pos\_rec}: coordinates perturbed with an alternative noise scale $\lambda$, serving as the reconstruction target; denoted by \( \bm{r}_i^{\texttt{rec}} \in \mathbb{R}^{3} \).

The injected noise \( \boldsymbol{\epsilon} \sim \mathcal{N}(0, \sigma^2) \) is shared across these variants. Specifically, we define:
\[
    \bm{r}_i^{\texttt{nsd}} = \bm{r}_i^{\texttt{cln}} + \boldsymbol{\epsilon}, \quad
    \bm{r}_i^{\texttt{rec}} = \bm{r}_i^{\texttt{cln}} + \lambda\cdot \boldsymbol{\epsilon}
\]
where \( \lambda \) modulates the reconstruction difficulty, set to 1 for default.  Each target coordinate is paired with its corresponding atomic number $z_i$ to construct the initial atomic embeddings.

Node-level embeddings are denoted as\( \{\mathbf{h}_i^{(*)}\}_{i=1}^N \), where \( * \in \{\texttt{dns}, \texttt{rec}, \texttt{cln}\} \) indicates the input variant corresponding to different pretraining tasks. In contrast, the graph-level representation is denoted by \( \mathbf{g} \in \mathbb{R}^d \), where $d$ represents the dimensionality of the embedding space. 

The model is trained to predict noise vectors \( \hat{\bm{\epsilon}}_i^{(*)} \in \mathbb{R}^3 \) for each atom and task, where the supervision signal is the synthetic Gaussian noise applied during pretraining. 
We denote the loss terms by \( \mathcal{L}_{*} \), where the subscript indicates the task name, and use \( \lambda_{*} \) for the corresponding weighting coefficients.

\subsection{Model Architecture}
\textbf{Shared Equivariant Encoder}\quad
At the core of {GeoRecon} is an 8-layer SE(3)-equivariant transformer encoder adapted from TorchMD-Net~\citep{tholke2022torchmd}. The encoder operates on molecular graphs defined by atomic numbers \( \{z_i\} \) and spatial coordinates \( \{\bm{r}_i^*\} \).



\textbf{Denoising Head}\quad
The denoising head is designed to reverse the synthetic perturbation applied during pretraining. We deliberately follow the node-level denoising paradigm to keep consistency to previous work and decrease migration difficulty, but our objectives explicitly enforce utilizing global graph structure. 

According to \autoref{theo-1}, the equivalence between coordinate denoising and force field learning under the Boltzmann distribution assumption can be readily established. A brief derivation is included in the appendix \ref{sec-pro} for completeness and reference. 

To fully leverage the clean molecular conformations already processed during pretraining, we integrate them into the denoising pipeline as well. Specifically, clean molecules, with zero perturbation, are passed through the same encoder and denoising head. Their supervision target is a zero noise vector, naturally enforcing representational consistency between clean and noised structures. This unifies the treatment of clean and noised molecules under a single objective, regularizing the encoder without requiring additional components or losses. To evaluate the impact of the auxiliary clean alignment task, we conduct additional ablation studies in \autoref{abl_clean}.

\textbf{Reconstruction Head with Graph-Level Conditioning}\quad
To mitigate the ambiguity introduced by global orientation, we design the reconstruction head differently from standard autoencoders. Instead of directly decoding absolute coordinates, GeoRecon formulates reconstruction as a more challenging denoising task: stronger noise is applied to the inputs, and the model predicts the corresponding perturbation with a lightweight decoder conditioned on the graph-level embedding. This design encourages the learning of structure-aware representations aligned with whole-molecule geometry, while leveraging global semantic information $\bm g$ derived from clean conformations to guide the denoising process. Intuitively, as the perturbation is very strong, the node embeddings after the encoder become extremely noisy. This time, if we directly concatenate the clean node embeddings with them, the decoder is easy to recover the added noise. Yet, we choose to not feed the clean node embeddings, but use the pooled clean graph embedding instead. This enforces the graph embedding to contain rich local information and be coherent with node-level representations. We make the decoder lightweight to further increase the task difficulty and thus reinforcing the reliance on the graph-level embeddings.

To construct the conditioning signal, the clean input undergoes a complete forward pass through the encoder, producing node-level representations \(\{ \bm{h}_i^{\texttt{(cln)}} \}\). These are aggregated via pooling function to yield a graph-level vector.
The resulting vector $\bm{g}$ encapsulates holistic structural semantics and is broadcasted to all nodes in the reconstruction input. We adopt mean pooling for its simplicity and stability under coordinate perturbations, while noting that more sophisticated pooling strategies are also compatible with our framework. Each node concatenates this global vector with its local representation \(\bm{h}_i^{(\texttt{rec})}\), forming the input to a lightweight decoder:
$$
\bm{z}_i = \textsc{Concat}(\bm{g}, \bm{h}_i^{\texttt{(rec)}}), \quad 
\hat{\boldsymbol{\epsilon}}_i^{\texttt{(rec)}} = \textsc{MLP}_{\texttt{rec}}(\bm{z}_i).
$$


Consequently, the learned latent space attains dual expressiveness: it remains \textit{locally descriptive, capturing fine-grained atomic perturbations}, while also being \textit{globally coherent in reflecting the overall molecular topology}. This property improves generalization in downstream molecular property prediction, especially for tasks that depend strongly on spatial organization, including total energy, dipole moment, and orbital energy estimation.

\subsection{Multi-Task Training Objective}

To jointly optimize local and global aspects of molecular geometry, we combine the three pretraining objectives into a unified loss:
\[
\mathcal{L}_{\texttt{total}} = \lambda_{\texttt{nsd}} \mathcal{L}_{\texttt{nsd}} + 
\lambda_{\texttt{rec}} \mathcal{L}_{\texttt{rec}}+ 
\lambda_{\texttt{cln}} \mathcal{L}_{\texttt{cln}}.
\]
Each loss term provides complementary supervision, encouraging the encoder to learn representations that are both locally descriptive and globally coherent. The loss weights are empirically chosen to balance the magnitudes of individual terms during training.

\section{Experiments}
Information about training devices, hyperparameters, and computational costs are provided in \autoref{app:t}. Training code is available in the supplementary materials.
\begin{figure*}[ht]
    \centering
    \vspace{-0.5em}
    \includegraphics[width=0.85\textwidth]{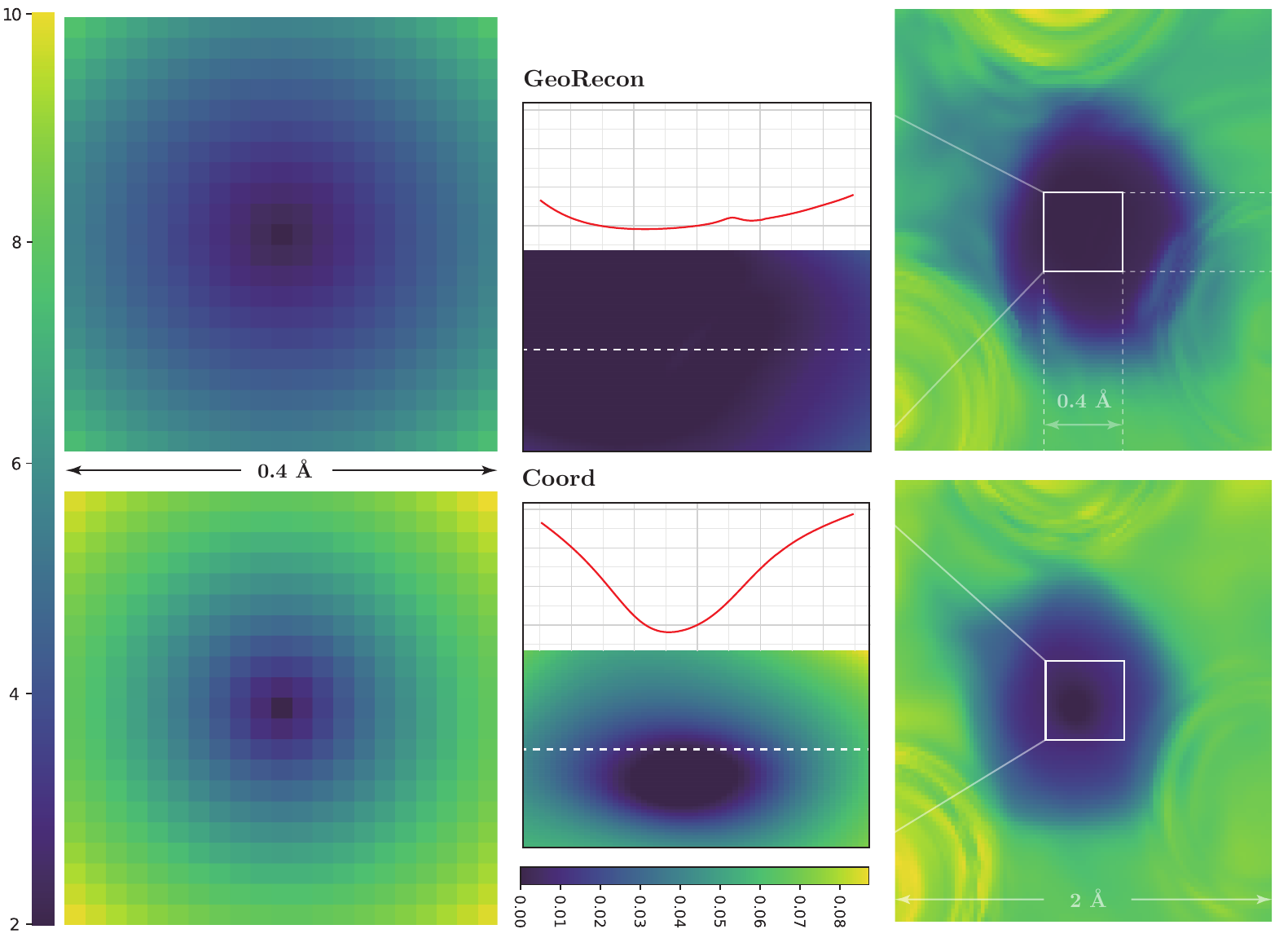}
    \caption{Representation stability of GeoRecon (upper row) vs. Coord (lower row) near equilibrium conformations ($\|\delta \bm{x}\| \leq 1$ \AA). 
    \textbf{Left}: averages over multiple molecules. 
    \textbf{Middle \& Right}: 2D perturbation heatmaps for a randomly selected PCQM4Mv2 sample, where the horizontal and vertical axes correspond to the magnitudes of two coordinate perturbations applied to a random atom, and the color encodes the norm change of the representation ($\|\delta\bm{h}\|$). Larger blue regions indicate higher stability of the representation near the equilibrium conformation. The red curve in the middle column highlights $\delta y=0$ ($\|\delta\bm{h}\|$ along the $x$-axis). 
    \textbf{Scale bars}: shared on the left (left column) and at the bottom (middle).}
    \vspace{-0.5em}
    \label{lip_fig}
\end{figure*}
\subsection{Lipshitz Constant Analyze after Pretraining}
We measured the Lipshitz constant of GeoRecon after pretraining on PCQM4Mv2 and observed a substantial reduction ($\sim$ 99\%) in GeoRecon compared to Coord.

As shown in \autoref{Lipshitz} , GeoRecon consistently achieves median $L(x)\approx$ 30 with negligible variance across step sizes, while the Coord baseline yields values on the order of $10^4\sim 10^5$, representing a reduction of roughly \emph{three orders of magnitude} in the local Lipschitz constant.

\autoref{lip_fig} compares the sensitivity of GeoRecon and Coord encoders near equilibrium conformations. Under small perturbations ($\|\delta \bm{x}\|\leq 0.4$ \AA), GeoRecon exhibits much smoother representation manifolds, with smaller changes in $\|\bm{h}\|$ compared to Coord. This trend is consistent both in individual samples and in averages across multiple molecules.

The experimental results confirm our intuition that anchoring a graph-level perspective during the pre-training stage leads to a smoother latent space. This provides a stronger theoretical foundation for our improvements in downstream tasks, as smoother representation spaces have been shown to facilitate better learning procedures in various other settings \citep{lee2025enhancing,guo2024smooth,zhang2025repcali,krishnan2020lipschitz}.

\subsection{Test on Downstream Tasks}

Our method is built upon the denoising task, with an additional reconstruction auxiliary task. Therefore, \emph{our primary direct comparison is with the denoising baseline \textbf{Coord}}~\citep{Zaidi2023}. To clearly highlight improvements over this baseline, we mark them in \textcolor{green!50!black}{green} in Tables~\ref{tab:qm9}, ~\ref{tab:md17} and~\ref{tab:md22_results1}. 

We include comparisons to traditional methods without pre-training to demonstrate the benefits of pretraining, as well as to several pre-training approaches. 

\begin{table*}[ht]
   
    \caption{MAE (↓) on QM9 property prediction. Best and second-best results are \textbf{bolded} and \underline{underlined}, respectively. Aver-rank is the mean of per-metric ranks. The methods are divided into two groups: training from scratch and pre-training then fine-tuning. Green text indicates the relative improvement (\%) over the base model; red indicates degradation.}
    \label{tab:qm9}
    \vspace{0.5em}
    \centering
        \resizebox{0.9\textwidth}{!}{
    \begin{tabular}{l|cccccccccccc|c}
    \toprule
     \textbf{Task}& $\mu$ & $\alpha$ & homo & lumo & gap & $R^2$ & ZPVE & $U_0$ & $U$ & $H$ & $G$ & $C_v$ & Average \\
     \emph{Unit}& \scriptsize(D) & \scriptsize($a_0^3$) & \scriptsize(meV) & \scriptsize(meV) & \scriptsize(meV) & ($a_0^2$) & \scriptsize(meV) & \scriptsize(meV) & \scriptsize(meV) & \scriptsize(meV) & \scriptsize(meV) & \scriptsize$\left(\tfrac{\text{cal}}{\text{mol}\cdot \text{K}}\right)$& rank\\
    \midrule
    SchNet        & 0.033 & 0.235 & 41.0 & 34.0 & 63.0 & 0.070 & 1.70 & 14.00 & 19.00 & 14.00 & 14.00 & 0.033 &9.33\\
    EGNN          & 0.029 & 0.071 & 29.0 & 25.0 & 48.0 & 0.106 & 1.55 & 11.00 & 12.00 & 12.00 & 12.00 & 0.031 &9.00 \\
    DimeNet++     & 0.030 & \underline{0.044} & 24.6 & 19.5 & 32.6 & 0.330 & 1.21 & 6.32  & 6.28  & 6.53  & 7.56  & 0.023 &6.84\\
    PaiNN         & \underline{0.012} & 0.045 & 27.6 & 20.4 & 45.7 & \underline{0.070} & 1.28 & \underline{5.85}  &\underline{5.83}  & \underline{5.98}  & 7.35  & 0.024 &5.67\\
    SphereNet     & 0.025 & 0.045 & 22.8 & 18.9 & \underline{31.1} & 0.270 & \textbf{1.12} & 6.26  & 6.36  & 6.33  & 7.78  & 0.022 &5.50 \\
    TorchMD-Net   & \textbf{0.011} & {0.059} & 20.3 & 17.5 & 36.1 & \textbf{0.033} & 1.84 & 6.15  & 6.38  & 6.16  & 7.62  & 0.026 & 4.17 \\
    \midrule
    Transformer-M & 0.037 & \textbf{0.041} & \underline{17.5} & 16.2 & \textbf{27.4} & 0.075 & \underline{1.18} & 9.37  & 9.41  & 9.39  & 9.63  & 0.022 &\underline{3.67}\\
    SE(3)-DDM     & 0.015 & 0.046 & 23.5 & 19.5 & 40.2 & 0.122 & 1.31 & 6.92  & 6.99  & 7.09  & 7.65  & 0.024 &6.33\\
    3D-EMGP       & 0.020 & 0.057 & 21.3 & 18.2 & 37.1 & 0.092 & 1.38 & 8.60  & 8.60  & 8.70  & 9.30  & 0.026 &6.00\\
    \rowcolor{gray!10}Coord         & 0.016 & 0.052 & 17.7 & \underline{14.7} & 31.8 & 0.450 & 1.71 & 6.57  & 6.11  & 6.45  & \underline{6.91}  & \textbf{0.020}&5.33 \\
    \rowcolor{gray!10}
    GeoRecon (Ours)  
    & \underline{0.012} & {0.048} & \textbf{16.9} & \textbf{14.2} & 31.4 & 0.238 & 1.51 & \textbf{5.15} & \textbf{5.09} & \textbf{5.13} & \textbf{6.39} & 0.022 & \textbf{3.50} \\
    \midrule
    \multicolumn{1}{l|}{\textbf{Relative Gain}} 
    & \textcolor{green!50!black}{\small 23.4\%} & \textcolor{green!50!black}{\small 7.0\%} & \textcolor{green!50!black}{\small 4.8\%} & \textcolor{green!50!black}{\small 3.5\%} & \textcolor{green!50!black}{\small 1.3\%} & \textcolor{green!50!black}{\small 47.2\%} & \textcolor{green!50!black}{\small 11.9\%} & \textcolor{green!50!black}{\small 21.7\%} & \textcolor{green!50!black}{\small 16.7\%} & \textcolor{green!50!black}{\small 20.5\%} & \textcolor{green!50!black}{\small 7.5\%} & \textcolor{red!50!black}{\small -10.0\%} &  \\
    \bottomrule
    \end{tabular}
    }
    \end{table*}
    
\subsubsection{Evaluation on QM9}
\textbf{QM9} benchmark~\citep{ramakrishnan2014quantum} contains 134k stable small organic molecules (up to 9 heavy atoms: C, O, N, F, and H), each optimized at the B3LYP/6-31G(2df,p) level of DFT. It provides geometric, energetic, electronic, and thermodynamic properties; in this work, we focus on the 3D geometries and associated properties relevant for molecular representation learning.

    See \autoref{tab:qm9} for results.  The details of baselines are shown in \autoref{app:bsl}. As shown, GeoRecon achieves strong performance across a range of molecular property prediction tasks. It achieves the best performance across all baselines on multiple targets, including $U$, $U_0$, $H$, and $G$. Compared to our backbone Coord, GeoRecon also shows \textit{noticeable improvements} on several other key targets. 
\vspace{-0.5em}
\begin{table}[ht]
    \centering
    \caption{MAE (↓) of force prediction on the MD17 dataset (kcal/mol/Å). 
    Best results are \textbf{bolded}. }
    \label{tab:md17}
    \vspace{0.5em}
    \resizebox{0.49\textwidth}{!}{
    \begin{tabular}{l|cccccccc}
    \toprule
    \textbf{Molecule} & Aspirin & Benzene & Ethanol & Malonaldehyde \\
    \midrule
Coord         & 0.23299	&0.15052	&0.10909	&0.16218\\
\centering GeoRecon (Ours)    
& \textbf{0.21097} &\textbf{0.14734}	&\textbf{0.09771}	&\textbf{0.15705} \\
\midrule
\rowcolor{gray!0}    
\textbf{Relative Gain}& {\small \textcolor{green!50!black}{9.45\%}} 
& {\small \textcolor{green!50!black}{2.11\%}} 
& {\small \textcolor{green!50!black}{10.43\%}} 
& {\small \textcolor{green!50!black}{3.16\%}} 
\\\midrule
\textbf{Molecule} &  Naphthalene & Salicylic Acid & Toluene & Uracil\\
    \midrule
Coord         &0.06266	&0.13336	&0.06843	&0.09109\\
\centering GeoRecon (Ours)    
&\textbf{0.05755}	&\textbf{0.11996}	&\textbf{0.06064}	&\textbf{0.08716}  \\
\midrule
\rowcolor{gray!0}    
\textbf{Relative Gain}

& {\small \textcolor{green!50!black}{8.14\%}} 
& {\small \textcolor{green!50!black}{10.05\%}} 
& {\small \textcolor{green!50!black}{11.37\%}} 
& {\small \textcolor{green!50!black}{4.32\%}} 
\\
    \bottomrule
    \end{tabular}
    }
    \vspace{-1em}
\end{table}
\subsubsection{Evaluation on MD17}
\textbf{MD17} benchmark~\citep{chmiela2017machine} consists of \emph{ab initio} molecular dynamics (MD) trajectories for small molecules (e.g., ethanol, malonaldehyde, glycine) at $500~\mathrm{K}$, providing tens of thousands of DFT-computed total energies and atomic forces per molecule. It has become a widely used benchmark for assessing machine-learned force fields in the low-dimensional, gas-phase regime.

As shown in Table~\ref{tab:md17}, GeoRecon consistently improves upon the Coord baseline across all MD17 force prediction tasks. The results demonstrate that incorporating graph-level reconstruction yields clear gains over coordinate denoising alone. 
\subsubsection{Evaluation on MD22}
\textbf{MD22} benchmark~\citep{chmiela2023accurate} extends MD17 to larger and more complex supramolecular systems (42–370 atoms), including peptides, lipids, carbohydrates, DNA base pairs, and nanotube complexes. It provides \emph{ab initio} MD trajectories at $400$–$500~\mathrm{K}$ with a $1~\mathrm{fs}$ timestep, and energies and forces computed at the PBE+MBD level of DFT, posing a greater challenge for ML-based force fields due to its size, flexibility, and strong non-local interactions.

\begin{table}[ht]
    \centering
    \vspace{-0.8em}
    \caption{MAE (↓) of different models on the MD22 dataset. Energy (kcal), Force (kcal/$\mathring{\rm A}$). Best results are \textbf{bolded}.}
    \vspace{0.5em}
    \resizebox{0.5\textwidth}{!}{
        \begin{tabular}{l|cccccc}
            \toprule
            \textbf{Molecule} & \multicolumn{2}{c}{DHA} & \multicolumn{2}{c}{Stachyose} & \multicolumn{2}{c}{AT-AT-CG-CG} \\
            \cmidrule(lr){2-3}\cmidrule(lr){4-5}\cmidrule(lr){6-7}
            \emph{Metric} & Energy & Force & Energy & Force & Energy & Force \\
            \midrule
        Coord         & 0.15772 & 0.13891 & 0.70198 & 0.52737 & 0.48079 & 0.34858 \\
        \centering GeoRecon   
        & \textbf{0.11955} & \textbf{0.13553} & \textbf{0.59882} & \textbf{0.51666} & \textbf{0.47458} & \textbf{0.34080}  \\
        \midrule
        \rowcolor{gray!0}    
        \textbf{Relative Gain} 
        & {\small \textcolor{green!50!black}{24.20\%}} 
        & {\small \textcolor{green!50!black}{2.43\%}} 
        & {\small \textcolor{green!50!black}{14.70\%}} 
        & {\small \textcolor{green!50!black}{2.03\%}} 
        & {\small \textcolor{green!50!black}{1.29\%}} 
        & {\small \textcolor{green!50!black}{2.23\%}} \\
            \bottomrule
        \end{tabular}
    }
    
    \label{tab:md22_results1}
    \end{table}
    \begin{figure*}[t]
    \centering
    \vspace{-0.65em}
    \includegraphics[width=0.9\textwidth]{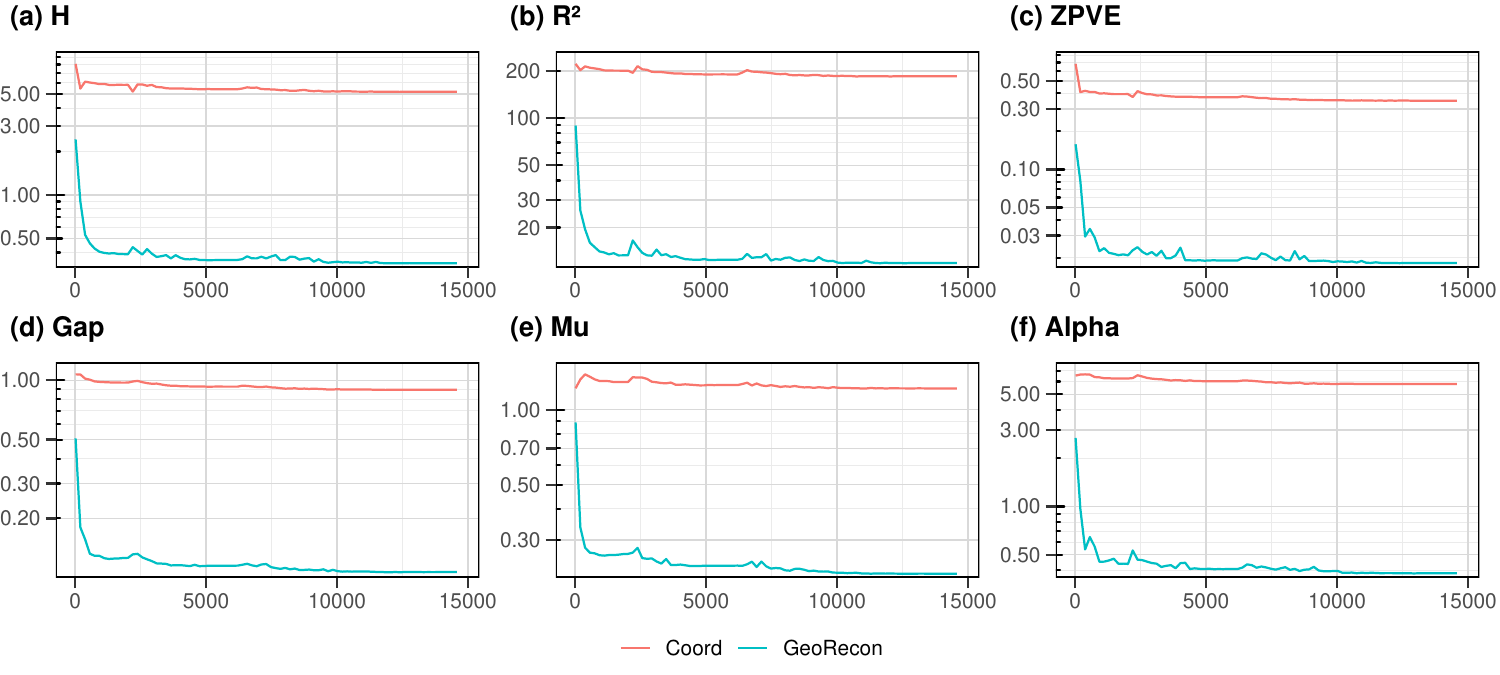}
    \vspace{-1em}
    \caption{MAE ($\downarrow$) loss curves from linear probing experiments on multiple QM9 tasks. A cosine–warmup learning rate schedule is applied, and both models are trained for 14.7k steps.}
    \label{fig:lp}
\end{figure*}\vspace{-0em}
We selected three molecules from the MD22 dataset: {DHA (56 atoms)}, {Stachyose (87 atoms)}, and {AT-AT-CG-CG (118 atoms)} to evaluate our method on small-, medium-, and large-scale systems, respectively. As shown in \autoref{tab:md22_results1}, GeoRecon outperforms the Coord backbone on both energy and force prediction tasks across all three molecules, demonstrating the effectiveness of our approach beyond small-molecule scenarios.

\vspace{-0.5em}
    

\subsection{Linear Probing Experiments}
In previous work~\citep{kumar2022fine}, it has been shown that linear probing, that is, appending a lightweight classification head to a frozen, well-pretrained representation model, can achieve stronger performance under out-of-distribution (OOD) settings. Motivated by this finding, we conduct linear probing experiments on GeoRecon and Coord to directly evaluate the influence of different pretraining methods. Since the encoder is kept frozen during these experiments, the degrees of freedom for training are significantly reduced. Consequently, the simplicity of the linear head imposes a strong requirement on the encoder to have already learned sufficiently good features during pretraining.

The experimental results are shown in \autoref{fig:lp}. GeoRecon consistently outperforms Coord across all tasks under the linear probing setting. Notably, Coord exhibits almost no optimization in this setting, suggesting that it can hardly adapt to downstream tasks without full fine-tuning that alters the pretrained structure. As discussed earlier, we attribute this to the limited quality of its pretrained representations. This observation is consistent with the Lipschitz constant measurements reported in the Introduction, further indicating that GeoRecon learns smoother and more informative representations.


\section{Conclution and Limitation}
Our work presents GeoRecon, a reconstruction-based pretraining framework that bridges node-level denoising and graph-level supervision for 3D molecular representation learning. By conditioning geometry reconstruction on global molecular embeddings, GeoRecon compels the encoder to capture both fine-grained atomic environments and coherent whole-molecule topology. Our theoretical analysis shows that this design produces substantially smoother latent manifolds—reducing local Lipschitz constants by nearly three orders of magnitude compared with node-centric baselines—which provably tightens generalization bounds and improves robustness to structural noise.

Comprehensive experiments across QM9, MD17, and MD22 validate the effectiveness of GeoRecon. The framework achieves competitive results on small-molecule property prediction and consistently improves upon the direct baseline Coord on molecular dynamics prediction tasks (MD17 and MD22) as well as on most targets in QM9. Ablation studies further show that reconstruction noise scaling and decoder depth govern the trade-off between local fidelity and global coherence.

Key advantages of GeoRecon include eliminating dependence on 2D graphs or external labels, seamless integration with SE(3)-equivariant architectures, and a principled connection between representation smoothness and downstream generalization. Looking forward, extending GeoRecon to biomacromolecules, contrastive or generative graph-level objectives, and broader biochemical benchmarks will further test its versatility. Taken together, these results establish geometry-aware graph-level reconstruction as a general paradigm for molecular pretraining, offering a simple yet powerful path toward more sample-efficient and physically grounded representation learning.

\bibliography{reference}
\bibliographystyle{iclr2025_conference}

\appendix
\onecolumn
\section{Ablation Studies}

\subsection{Ablation of Task Rec and Clean}

To assess the contributions of different hyperparameters in our GeoRecon framework, we conduct ablation studies along three axes: decoder depth $L$, reconstruction noise scale $\lambda$, and reconstruction loss weight $\lambda_{\texttt{rec}}$.

\paragraph{Effect of Reconstruction Noise Scale $\lambda$.}
In \autoref{notations} we pointed out the influence of the reconstruction noise scale $\lambda$, which directly modulates the difficulty of the pretraining task. 
To further validate our conclusions, we performed additional experiments in response to Reviewer CY2D’s comments. 
\autoref{A.1.5} reports both the measured Lipschitz constants of the learned representations and the downstream performance on the QM9 $H_{\text{atom}}$ task, using a 14-layer encoder pretrained on a 10k subset of QM9 with varying $\lambda$. 

\begin{table}[h]
\centering
\caption{Effect of reconstruction noise scale $\lambda$ on Lipschitz constant (Lip) and downstream MAE ($\downarrow$) for the $H_{\text{atom}}$ task. Pretraining is performed on a 10k subset of QM9 with a 14-layer encoder.}
\label{A.1.5}
\vspace{0.5em}\resizebox{0.45\textwidth}{!}{
\begin{tabular}{c|cccc}
\toprule
$\lambda$ & 1.15 & 1.20 & 1.35 & 1.50 \\
\midrule
Lip       & 64.505 & 59.934 & 59.489 & 62.945 \\
$H_{\text{atom}}$ & 7.0172 & 6.5091 & \textbf{6.4686} & 6.6525 \\
\bottomrule
\end{tabular}}
\end{table}

As shown, moderate values of $\lambda$ (e.g., $1.20$–$1.35$) yield both reduced Lipschitz constants and improved downstream accuracy, suggesting smoother and more transferable representations. 
This observation aligns with our intuition: larger $\lambda$ injects stronger noise, making reconstruction more challenging and forcing the model to capture global molecular dependencies, whereas excessively large $\lambda$ degrades stability. 
Notably, the $\lambda$ values used in our main experiments were not exhaustively tuned, implying that additional calibration may further enhance GeoRecon’s performance. 

\textbf{Effect of Decoder Depth and Noise Scale.} We vary the number of layers $L \in \{3,4,5\}$ in the lightweight decoder and the reconstruction noise scale $\lambda \in \{1.0, 1.5\}$, which controls the magnitude of perturbation in the reconstruction targets. 

\begin{wraptable}[15]{r}{0.52\textwidth}
    \centering
    \vspace{-2em}
    \caption{Effect of decoder depth $L$ and noise scale $\lambda$ on finetuning performance (MAE$\downarrow$). Each group reports HOMO, LUMO, and GAP errors, the best result in each task is highlighted in \textbf{bold}, and the second-best is \underline{underlined}, respectively. (Pre-training 10 epoch)}
    \label{tab:wrap-depth}
    \vspace{0.5em}
    \resizebox{0.48\textwidth}{!}{
    \begin{tabular}{c|c|c|c|c}
    \toprule
    $\lambda$ & Task & $L=3$ & $L=4$ & $L=5$ \\
    \midrule
    \multirow{3}{*}{1.0} & HOMO & 0.01812 & 0.02028 & \underline{0.01687} \\
                         & LUMO & 0.01506 & 0.02110 &\textbf{ 0.01368} \\
                         & GAP  & \textbf{0.03132} & 0.04220 & \underline{0.03157} \\
    \midrule
    \multirow{3}{*}{1.5} & HOMO & 0.01722 & \textbf{0.01661} & 0.01688 \\
                         & LUMO & \underline{0.01472} & 0.01451 & 0.01497 \\
                         & GAP  & 0.03218 & 0.03307 & 0.03238 \\
    \bottomrule
    
    \end{tabular}
    }
    
\end{wraptable}

The experiments results are shown in Table~\ref{tab:wrap-depth}. These results suggest an interaction between decoder depth and noise scale. When the lightweight decoder is shallow (e.g., \( L = 3 \) or \( L = 4 \)), increasing the noise scale from \( \lambda = 1.0 \) to \( \lambda = 1.5 \) improves downstream performance, likely because stronger perturbations force the encoder to rely more heavily on the global representation to support accurate reconstruction.

However, when the decoder becomes deeper (\( L = 5 \)), it gains sufficient capacity to locally denoise without depending as much on the global context. As a result, increasing the noise scale may not bring further benefit, and can even lead to performance degradation due to a mismatch between input difficulty and the model's reliance on global structure. The results suggest that shallower decoders rely more on graph-level guidance, supporting our core hypothesis about the need for global conditioning.

\textbf{Effect of Reconstruction Loss Weight.} In our ablation study with 30-epoch pretraining and a CosineWarmup schedule, we varied the reconstruction loss weight \(\lambda_{\texttt{rec}} \in \{0.40, 0.45, 0.50\}\) while fixing decoder depth \(L=5\) and noise scale \(\lambda=1.0\). 

The results in Table~\ref{tab:wrap-ratio} reveal a non-monotonic trend as $\lambda_{\texttt{rec}}$ varies. A low weight (0.40) leads to better performance on HOMO, while higher values (e.g., 0.50) improve LUMO and gap. The intermediate setting ($\lambda_{\texttt{rec}} = 0.45$) achieves the most balanced performance. The results suggest that $\lambda_{\texttt{rec}}$ modulates the encoder’s representational focus. We interpret this as an empirical observation that stronger reconstruction constraints may influence the encoder’s global focus, motivating further analysis.
\begin{table}[h]
    \centering
    \vspace{-1em}
    \caption{Effect of reconstruction loss weight $\lambda_{\texttt{rec}}$ with fixed $L=5$, $\lambda=1$, the best result in each task is highlighted in \textbf{bold}. (Pre-training 30 epoch)}
    \label{tab:wrap-ratio}
    \vspace{0.75em}
    \resizebox{0.38\textwidth}{!}{
    \begin{tabular}{c|ccc}
    \toprule
    $\lambda_{\texttt{rec}}$ & HOMO & LUMO & GAP \\
    \midrule
    0.4 & \textbf{0.01630} & 0.01509 & 0.03345 \\
    0.45 & 0.01756      & \textbf{0.01379}       & \textbf{0.03141} \\
    0.5 & 0.01659 & 0.01403 & 0.03227 \\
    \bottomrule
    \end{tabular}}
    \vspace{-10pt}
\end{table}

\section{Method of Lipshitz Constant Analyze}
\label{app:b}
We measured the spectral norm of the Jacobian PCQM4Mv2 for both GeoRecon and Coord after the same pre-training period, and defined the \textit{Local Lipschitz Constant} (LLC) at a conformation $x$ as follows \citep{miyato2018spectral}:
\begin{equation*}
    L(x) = \|J_f(x) P\|_2
\end{equation*}
Here, $f$ denotes the encoder, and $J_f(x) \in \mathbb{R}^{d \times 3N}$ is the Jacobian of $f$ at $x$. The orthogonal projector $P \in \mathbb{R}^{3N\times 3N}$ onto the non‑rigid subspace is applied to remove rigid‑body degrees of freedom, so that the measured local Lipschitz constant reflects sensitivity only to physically meaningful (non‑rigid) deformations and is independent of global SE(3) motions. Following \citet{gouk2021regularisation} and \citet{novak2018sensitivity}, we approximate $L(x)$ via power iteration.
\section{Why Smoother Representations Improve Finetuning}
\label{sec:smooth_finetune}

\paragraph{Setting.}
Let $(\mathcal X,\bar d)$ denote the space of molecular conformations modulo rigid motions, where $\bar d$ is the Procrustes / RMSD distance
\begin{equation*}
\bar d(x,x') \;:=\; \min_{R\in \mathrm{SO}(3),\,t\in\mathbb R^3} \big\|x - (Rx'+t)\big\|_2.
\end{equation*}
Let $f:(\mathcal X,\bar d)\!\to\!(\mathbb R^m,\|\cdot\|_2)$ be a \emph{graph-level} representation map with Lipschitz constant $L_f$, i.e.,
$\|f(x)-f(x')\|_2 \le L_f \,\bar d(x,x')$.
We assume a training sample $S=\{(x_i,y_i)\}_{i=1}^n$ with diameter $\mathrm{diam}_{\bar d}(S)\le D$.
The downstream loss $\ell$ is assumed to be $1$-Lipschitz in its prediction argument and bounded (or rescaled to $[0,1]$ for concentration).%
\footnote{Boundedness is only used for the empirical-to-population concentration term.}
All results are stated on the SE(3)-\emph{quotient} space and thus respect molecular symmetries.

\subsection{Linear-Probe Analysis (Frozen Encoder)}
We consider a linear probe on top of the frozen $f$:
$g_w(z)=\langle w,z\rangle+b$ with $\|w\|_2\le B$; denote $\mathcal H_{\mathrm{lin}}=\{x\mapsto g_w(f(x))\}$.

\begin{theorem}[Smoother representations tighten generalization for linear probes]
\label{thm:lin_gen}
With probability at least $1-\delta$, every $h\in\mathcal H_{\mathrm{lin}}$ satisfies
\begin{equation*}
\mathcal R(h) \;\le\; \widehat{\mathcal R}_S(h) \;+\; 
\frac{c\, B\, L_f\, D}{\sqrt{n}} \;+\;
\sqrt{\frac{\log(1/\delta)}{2n}},
\label{eq:lin_bound}
\end{equation*}
for an absolute constant $c$.
\end{theorem}

\begin{proof}[Sketch]
Apply the contraction lemma to $\phi_y(u)=\ell(y,u)-\ell(y,0)$, which is $1$-Lipschitz and satisfies $\phi_y(0)=0$, to obtain
$\mathfrak R_n(\ell\!\circ\!\mathcal H_{\mathrm{lin}})\le \mathfrak R_n(\mathcal H_{\mathrm{lin}})$.
For $\mathcal H_{\mathrm{lin}}$ we have
\[
\mathfrak R_n(\mathcal H_{\mathrm{lin}})\le \frac{B}{n}\,\mathbb E_\sigma\Big\|\sum_{i=1}^n \sigma_i f(x_i)\Big\|_2.
\]
Pick any reference $x_0$ and set $c=f(x_0)$; the bias $b$ absorbs this shift. Since
$\|f(x_i)-c\|_2\le L_f\,\bar d(x_i,x_0)\le L_f D$ for all $i$,
\[
\mathbb E_\sigma\Big\|\sum_{i=1}^n \sigma_i\big(f(x_i)-c\big)\Big\|_2
\;\le\;\Big(\sum_{i=1}^n \|f(x_i)-c\|_2^2\Big)^{1/2}
\;\le\;\sqrt{n}\,L_f D.
\]
Hence $\mathfrak R_n(\mathcal H_{\mathrm{lin}})\le (B L_f D)/\sqrt{n}$.
Plugging this into the standard Rademacher generalization bound for bounded $1$-Lipschitz losses yields \eqref{eq:lin_bound}.
\end{proof}

\begin{lemma}[Noise robustness scaling with $L_f$ under a linear probe]
\label{lem:lin_noise}
Let the input be perturbed by additive Gaussian noise $x\mapsto x+\varepsilon$ with $\varepsilon\sim \mathcal N(0,\sigma^2 I)$ in Cartesian coordinates (for $N$ atoms hence $3N$ dimensions).
Then for any $w$ with $\|w\|_2\le B$,
\begin{align*}
\mathbb E_\varepsilon\!\left[\,\big|g_w(f(x+\varepsilon)) - g_w(f(x))\big|\,\right]
&\le B\,L_f\,\mathbb E_\varepsilon\|\varepsilon\|_2 = \Theta\big(B\,L_f\,\sigma\sqrt{3N}\big), \\
\mathbb E_\varepsilon\!\left[\,\big(g_w(f(x+\varepsilon)) - g_w(f(x))\big)^2\,\right]
&\le B^2 L_f^2\,\mathbb E_\varepsilon\|\varepsilon\|_2^2 = 3N\sigma^2 B^2 L_f^2.
\end{align*}
\end{lemma}

\begin{proof}[Sketch]
Alignment only decreases distance:
\[
\bar d(x,x+\varepsilon)=\min_{R,t}\|x-(R(x+\varepsilon)+t)\|_2
\le \|x-(x+\varepsilon)\|_2=\|\varepsilon\|_2.
\]
Thus,
\[
|g_w(f(x+\varepsilon))-g_w(f(x))|
\le \|w\|_2\,\|f(x+\varepsilon)-f(x)\|_2
\le B L_f \bar d(x,x+\varepsilon)\le B L_f \|\varepsilon\|_2.
\]
Taking (squared) expectations with respect to $\varepsilon$ gives the claims.
\end{proof}

The inequality $\bar d(x, x+\varepsilon) \le \|\varepsilon\|_2$ is a conservative upper bound, 
since the optimal alignment $(R,t)$ typically yields a much smaller distance in practice. 
Therefore, our result should be interpreted as a worst-case guarantee: 
the amplification of input perturbations cannot exceed $B L_f \|\varepsilon\|_2$, 
while in realistic settings the effect is usually weaker.

\paragraph{Implication.}
Equations~\eqref{eq:lin_bound} and Lemma~\ref{lem:lin_noise} show that a smaller $L_f$ (i.e., a smoother $f$) strictly tightens linear-probe generalization and attenuates noise amplification.

\subsection{End-to-End Finetuning: Rigorous and Heuristic Routes}
Consider now an end-to-end predictor $h=g\circ f$ where $g:\mathbb R^m\to\mathbb R$ is a small readout network (e.g., the MLPs used in our experiments).
Let $L_g$ denote the Lipschitz constant of $g$ on the image of $f$ over the data domain.
The composite Lipschitz constant satisfies
\begin{equation}
L_h \;\le\; L_g  L_f .
\label{eq:composite}
\end{equation}

\paragraph{A rigorous route under spectral control.}
\begin{proposition}[Generalization and robustness with spectrally-controlled readouts]
\label{prop:full_spectral}
Let $g$ be a depth-$d$ MLP with $1$-Lipschitz activations (e.g., ReLU) and weight matrices $W_1,\dots,W_d$.
Assume spectral norm bounds $\|W_j\|_{\mathrm{op}}\le s_j$ and define the product of spectral norms $P := \prod_{j=1}^d s_j$ and the complexity factor
$S:=\big(\sum_{j=1}^d \|W_j\|_F^2/s_j^2\big)^{1/2}$.
Let $L_g\le P$ and $h=g\circ f$.
Then, with probability at least $1-\delta$,
\begin{equation*}
\mathcal R(h) \,\le\, \widehat{\mathcal R}_S(h) \,+\, 
\tilde c\frac{PSL_fD}{\sqrt{n}} \,+\,
\sqrt{\frac{\log(1/\delta)}{2n}},
\qquad
\mathbb E_\varepsilon\!\left[\,(h(x+\varepsilon)-h(x))^2\,\right]
\,\le\, (L_g L_f)^2 \cdot 3N\sigma^2,
\label{eq:full_bounds}
\end{equation*}
for a universal constant $\tilde c$ (independent of dimensions and model parameters).
\end{proposition}

\begin{proof}[Sketch]
By \eqref{eq:composite}, $h$ is $L_h$-Lipschitz with $L_h\le L_g L_f$; the noise bound follows as in Lemma~\ref{lem:lin_noise}.
For generalization, apply a vector (Gaussian) contraction inequality to the composed class:
\[
\mathfrak R_n(\{g\circ f\}) \;\le\; C\,\Big(\prod_{j=1}^d s_j\Big)\,
\frac{1}{n}\,\mathbb E\Big\|\sum_{i=1}^n \gamma_i f(x_i)\Big\|
\;\le\; C\,\Big(\prod_{j=1}^d s_j\Big)\, \frac{L_f D}{\sqrt n},
\]
and incorporate the standard spectral/Frobenius control via
$S=\big(\sum_{j=1}^d \|W_j\|_F^2/s_j^2\big)^{1/2}$ to obtain the stated
$\tilde c\,\frac{P S L_f D}{\sqrt{n}}$ (constants absorbed into $\tilde c$).
\end{proof}

\paragraph{Remark.}
In our experiments, we \emph{did not} explicitly impose spectral norm constraints via dedicated regularization. 
However, common practices such as weight decay and initialization schemes tend to implicitly control the spectral norms of the weights, preventing $L_g$ and the complexity factor $S$ from becoming excessively large. 
Thus Proposition~\autoref{prop:full_spectral} should be interpreted primarily as a conceptual tool, clarifying how representation smoothness ($L_f$) and readout complexity jointly govern generalization.

Beyond such constrained settings, we next discuss a heuristic justification based on early-stage linearization.

\paragraph{A heuristic route via early-stage linearization.}
While the exact Rademacher complexity of general MLPs depends on architectural details, early end-to-end training is well-approximated by a linearized model around the pretrained parameters (NTK-style local linearization).
Under this viewpoint, the estimation term scales as $\tilde{\mathcal O}(L_g L_f D/\sqrt{n})$ with $L_g$ the \emph{local} Lipschitz constant near initialization, and the noise bound remains governed solely by the composite Lipschitz constant $(L_g L_f)$.

To empirically support our heuristic argument on early-stage linearization, 
we conduct a sanity-check experiment measuring both the cosine similarity between the predictions of the full model and its NTK approximation, and the alignment of parameter gradients across steps. 
As shown in Table~\ref{tab:linearization_sanity}, both metrics remain high ($>0.85$) during the first 300 steps, confirming that finetuning dynamics are well-approximated by a linear regime in the early stage. 
Although the similarity gradually decreases later, this evidence substantiates our claim that smoother encoders (smaller $L_f$) can benefit optimization through improved linearized dynamics.

\begin{table}[h]
    \centering
    \caption{Sanity check for linearization regime during early finetuning. 
    We report the cosine similarity between predictions of the full model and its linearized NTK approximation, 
    as well as the gradient alignment, measured as the cosine similarity between parameter gradients at consecutive steps. 
    Higher values indicate stronger validity of the linearization heuristic.}\vspace{0.5em}
    \label{tab:linearization_sanity}\resizebox{0.55\textwidth}{!}{
    \begin{tabular}{lcc}
    \toprule
    Step Range & Cosine Sim. (Full vs. NTK) & Grad. Alignment \\
    \midrule
    0--100  & 0.98 & 1.000 \\
    100--200 & 0.90 & 0.927 \\
    200--300 & 0.87 & 0.921 \\
    \bottomrule
    \end{tabular}}
    \end{table}
    
    As expected, the similarity metrics are highest in the very first steps and gradually decrease, 
    which aligns with the known phenomenon that NTK approximations are most accurate during the initial phase of training.

    \paragraph{Practical relevance.}
    Two considerations explain why full finetuning can match or surpass linear probes:
    (i) \emph{Certificate existence:} the hypothesis class for full finetuning strictly contains that of linear probes (by taking $g$ to be linear and freezing $f$), so Theorem~\ref{thm:lin_gen} remains attainable as a special case. 
    (ii) \emph{Local linearization:} in early training, dynamics are well-approximated by a linearized model around the pretrained parameters, effectively optimizing a linear head on a fixed $f$ and thus retaining the same $L_f$ dependence; with spectral/weight decay control, $L_g$ (and hence $L_h$) stays finite. 
    If end-to-end finetuning underperforms compared to linear probing, this often suggests limitations in the optimization procedure’s ability to leverage the expanded hypothesis class, rather than deficiencies in the representation itself.

\paragraph{Conclusion.}
Both the linear-probe guarantee (Theorem~\ref{thm:lin_gen}) and the spectrally-controlled end-to-end bound (Proposition~\ref{prop:full_spectral}), together with our empirical ablations, indicate that pretraining strategies which produce \emph{smoother} graph-level encoders (smaller $L_f$) can improve finetuning generalization and reduce sensitivity to coordinate perturbations. 
Practically, spectral or norm-based regularization on the readout complements smaller $L_f$, yielding tighter overall control via $L_h \le L_g L_f$.

\paragraph{Scope of analysis.} 
Our theoretical results should be interpreted at three different levels of rigor. 
(i) Theorem~\ref{thm:lin_gen} provides a tight and formally complete guarantee for the case of \emph{linear probing}, where the dependence on $L_f$ can be isolated exactly. 
(ii) Proposition~\ref{prop:full_spectral} extends this guarantee to nonlinear readouts under explicit spectral norm and Frobenius norm control, which mathematically preserves the same $L_f$-dependence but requires additional architectural constraints. 
(iii) Beyond these constrained settings, inspired by \citet{jacot2018neural}, we provide a heuristic justification via early-stage linearization, suggesting that the benefits of smoothness $L_f$ are likely to extend to practical end-to-end finetuning. 
While the fully unconstrained nonlinear case remains an open theoretical question, our experiments in main content and \autoref{A.1.5} consistently support the practical value of smoother representations across both linear and nonlinear downstream models.

\paragraph{Limitation.} 
Our guarantees are rigorous for linear probes; extensions to nonlinear readouts rely on either architectural constraints or heuristic arguments, and the fully unconstrained case remains open. 
Nonetheless, our experiments consistently indicate that smoother representations ($L_f$) improve performance across both linear and nonlinear settings.

\section{Proof of equivariance between denoise and force field learning}

We briefly revisit the theoretical grounding of denoising as an approximation to physical force learning, as established by prior work~\citep{Zaidi2023}.

Let $\bm{x}_{i(j)}$ denote the position of atom $i$ in molecule $j$, where each conformation $\bm{x}_i \in \mathbb{R}^{3N}$ is mean-centered to ensure $\sum_j \bm{x}_{i(j)} = \bm{0}$, thereby lying in a $(3N - 3)$-dimensional subspace. Denote the potential energy function as $U(\bm{x})$, which maps geometric configurations to their corresponding energy levels. The negative gradient $-\nabla_{\bm{x}} U(\bm{x})$ corresponds to the force field and serves as the prediction target in force-learning tasks. The distribution of molecular conformations $\bm{x}$ is denoted as $p_{\text{phy}}(\bm{x})$, which follows the Boltzmann distribution and takes the form $A\exp\left(-U(\bm x)/kT\right)$ where $A$ is a normalization constant, $k$ is the Boltzmann constant, and $T$ is the temperature.

From a probabilistic perspective, the distribution of the noised coordinates $\bm{x}^{\texttt{nsd}}$ can be expressed as a marginalization over the clean coordinates:
 \begin{equation*}
        p(\bm x^{\texttt{nsd}})=\int p_{\tau}(\bm x^{\texttt{nsd}}|\bm x^{\texttt{cln}})p(\bm x^{\texttt{cln}
        })\,\mathrm{d} \bm x^\texttt{cln}
 \end{equation*}
 
 Here, the conditional distribution $p_\tau(\bm{x}^{\texttt{nsd}} \mid \bm{x}^{\texttt{cln}})$ is modeled as a Gaussian in the $(3N - 3)$-dimensional mean-centered subspace, with isotropic variance $\tau^2$. According to \citet{vincent2011connection}, {Denoising Score Match (DSM)} is equivalant to  Explicit Score Matching (ESM):
 \begin{align*}
    \mathcal L_{\mathrm{DSM}}&=\mathbb E_{p_\tau(\bm x^{\texttt{nsd}},\bm x^{\texttt{cln}})}\|\mathrm{GNN}_{\theta}(\bm x^{\texttt{nsd}})-\frac{\partial\log p_{\tau}}{\partial \bm x^{\texttt{nsd}}}\|^2\\ &=\mathbb E_{p_\tau(\bm x^{\texttt{nsd}},\bm x^{\texttt{cln}})}\|\mathrm{GNN}_{\theta}(\bm x^{\texttt{nsd}})-\frac 1{\tau^2}(\bm x^{\texttt{nsd}}-\bm x^{\texttt{cln}})\|^2\\
    &=\mathbb E_{p_{\mathrm{phy}}(\bm x^{\texttt{nsd}})}\|\mathrm{GNN}_{\theta}(\bm x^{\texttt{nsd}})-(-\nabla U(\bm x^{\texttt{nsd}}))\|^2+\mathrm{Const}=\mathcal{L}_{\mathrm{ESM}}+\mathrm{Const}
 \end{align*}

 This establishes a theoretical bridge between denoising-based supervision and force field learning under equilibrium statistics. Accordingly, our reconstruction task, which leverages graph-level embeddings to modulate scaling denoising, is not only empirically effective but also theoretically justified through its approximation of physically meaningful gradients. This connection underlies our decision to instantiate reconstruction via a scaling denoise mechanism.

 This interpretation is also supported by the general connection between denoising and score matching established in~\citet{vincent2011connection}.
\label{sec-pro}
\section{Ablation study on task clean and rec}
\label{abl_clean}

To rigorously demonstrate the contributions of our reconstruction task and the clean alignment task, we conducted an additional ablation study on them. We pretrained a RecOnly Model, which set the weight of Clean task weight into 0 and other hyperparameters keep unchanges. 

\begin{table}[h]
    \centering
    \vspace{-1em}
    \caption{Effect of clean alignment (MAE$\downarrow$) with fixed $L=5$, $\lambda=1$, under three pre-training settings: {Baseline}, {RecOnly}, and {Rec+Clean}. The best result for each target is highlighted in \textbf{bold}. (Pre-training for 30 epochs)}
    
    \label{tab9}
    \vspace{0.75em}
    \resizebox{0.40\textwidth}{!}{
    \begin{tabular}{c|ccc}
    \toprule
    Model & HOMO & LUMO & GAP \\
    \midrule
    Coord & 0.0177 & 0.0147 & 0.0318 \\
    RecOnly & \textbf{0.0163}      & 0.0141       & {0.0315} \\
    Rec+Clean & {0.0166} & \textbf{0.0137} & \textbf{0.0306} \\
    \bottomrule
    \end{tabular}}
    \vspace{-10pt}
\end{table}

The experimental results are shown in Table~\ref{tab9}. Our method (Rec+Clean) achieves the best performance in most of the three settings, which demonstrates the effectiveness of our design. Besides, RecOnly performs better than Coord, while the lack of the Clean task leads to slight performance drop compared to Rec+Clean, suggesting that the Clean task also contributes to the overall performance.

\vspace{1.5em}

\section{Evaluation on 3BPA}
\textbf{3BPA} dataset~\citep{musaelian2023learning} consists of \emph{ab initio} MD trajectories for the flexible drug-like molecule 3-(benzyloxy)pyridin-2-amine (3BPA), characterized by three central rotatable dihedral angles that yield a complex potential energy surface with multiple local minima. Configurations were generated via MD simulations at $300$, $600$, and $1200~\mathrm{K}$ (25~ps length, $1~\mathrm{fs}$ timestep, Langevin thermostat) and re-evaluated using DFT ($\omega$B97X/6-31G(d)). This dataset provides an explicit out-of-distribution (OOD) scenario absent in equilibrium datasets such as QM9 and MD17, making it a stringent testbed for generalization.

To assess the robustness of GeoRecon, we pretrained and finetuned both GeoRecon and its backbone model (TorchMD-Net) on 3BPA for 100 epochs under identical hyperparameters. Training was performed using conformations sampled at $300~\mathrm{K}$, while evaluation covered conformations at $300$, $600$, and $1200~\mathrm{K}$. Results are summarized in Table~\ref{tab:3bpa}.

\begin{wraptable}{r}{0.6\linewidth}
\centering
\vspace{-1em}
\caption{MAE ($\downarrow$) on the 3BPA dataset at different temperatures.}
\vspace{0.5em}
\label{tab:3bpa}\resizebox{0.55\textwidth}{!}{
\begin{tabular}{lcccc}
\toprule
Method & Metric & 300K & 600K & 1200K \\
\midrule
\multirow{2}{*}{TorchMD-Net} 
 & Energy & 0.08606 & 0.08734 & 0.16689 \\
 & Force  & 0.19291 & 0.19935 & 0.23835 \\
\midrule
\multirow{2}{*}{GeoRecon (Ours)} 
 & Energy & \textbf{0.04646} & \textbf{0.07252} & \textbf{0.12662} \\
 & Force  & \textbf{0.15367} & \textbf{0.17117} & \textbf{0.22662} \\
\bottomrule
\end{tabular}}
\end{wraptable}

GeoRecon achieves lower energy and force errors than its backbone across all temperatures, demonstrating improved stability under distribution shift. These results confirm that reconstruction-based pretraining enhances generalization from small equilibrium molecules to complex, flexible drug-like systems, highlighting the broader applicability of GeoRecon beyond standard benchmarks.
\section{Transferability of GeoRecon}
To further examine the transferability of GeoRecon, we conduct a reconstruction-pretraining experiment based on UniMol. 
We compare the performance of the reconstructed UniMol model ({Unimol\_Rec}) against a faithfully reproduced baseline ({Unimol (Reproduced)}) 
across five widely used benchmarks: 
\textbf{FreeSolv}~\citep{mobley2014freesolv} provides hydration free energies for small molecules, reflecting solvation effects in aqueous environments. 
\textbf{Lipo}~\cite{wu2018moleculenet} contains experimentally measured octanol–water partition coefficients ($\log P$), a key indicator of molecular permeability and bioavailability in drug discovery. 
\textbf{QM7}~\citep{blum,rupp} includes atomization energies of 7,165 molecules with up to 23 atoms, computed using density functional theory (DFT), and serves as an early benchmark for energy prediction. 
\textbf{QM8}~\citep{ruddigkeit2012enumeration,ramakrishnan2015electronic} reports electronic spectra and excited-state properties of $\sim$22k molecules, testing the capacity of models to capture quantum phenomena beyond ground-state energies. 
For \textbf{QM9}~\citep{ramakrishnan2014quantum}, we follow the UniMol setting and report results on the HOMO–LUMO gap prediction task.

\begin{table}[h]
    \centering
    \vspace{-1em}
    \caption{MAE ($\downarrow$) comparison between Unimol (Reproduced) and Unimol\_Rec.}\vspace{0.5em}
    \resizebox{0.45\textwidth}{!}{\begin{tabular}{lcc}
    \toprule
    Dataset & Unimol (Reproduced) & Unimol\_Rec \\
    \midrule
    FreeSolv & 1.6869  & \textbf{1.6341} \\
    Lipo     & \textbf{0.6121}  & 0.6230 \\
    QM7      & 47.1637 & \textbf{44.7910} \\
    QM8      & 0.0157  & \textbf{0.0155} \\
    QM9      & 0.0046  & \textbf{0.0046} \\
    \bottomrule
    \end{tabular}}
\end{table}

Overall, {Unimol\_Rec} yields noticeable gains on FreeSolv and QM7, while maintaining competitive or slightly better performance on QM8 and QM9. 
These findings suggest that incorporating reconstruction pretraining improves robustness across tasks without introducing extra complexity, 
underscoring the adaptability of GeoRecon as a transferable paradigm within existing molecular learning frameworks.

\section{Effect of Pretraining Dataset Scale on Performance}
The size of the pretraining dataset is a critical factor influencing the effectiveness of self-supervised molecular representation learning. To investigate the scalability of GeoRecon, we conducted controlled experiments by pretraining from scratch on two subsets of the PCQM4Mv2 dataset containing $10$k and $100$k molecules, respectively. After pretraining, the models were finetuned on the QM9 enthalpy prediction task. 

We systematically varied the encoder depth ($L=8,10,12$) and the reconstruction noise scaling factor $\lambda \in \{1.00, 1.05\}$, and report the mean absolute error (MAE) in Table~\ref{tab:scale}. 

\begin{table}[h]
\centering
\caption{MAE ($\downarrow$) on QM9 enthalpy(H) after pretraining GeoRecon on PCQM4Mv2 subsets of different sizes.}
\label{tab:scale}\resizebox{0.65\textwidth}{!}{
\begin{tabular}{c|ccc|ccc}
\toprule
\multirow{2}{*}{$\lambda$ / Encoder} & \multicolumn{3}{c|}{10k Subset} & \multicolumn{3}{c}{100k Subset} \\
 & \small $L=8$ & \small$L=10$ & \small$L=12$ & \small$L=8$ & \small$L=10$ & \small$L=12$ \\
\midrule
1.00 & 8.438 & 7.074 & 7.572 & 6.109 & \textbf{5.623} & 6.220 \\
1.05 & –     & 7.276 & 7.360 & 6.608 & 5.936 & 6.089 \\
\bottomrule
\end{tabular}}
\end{table}

The results reveal two main observations.  
First, dataset size substantially impacts pretraining quality: models pretrained on the $100$k subset achieve consistently lower errors than those trained on the $10$k subset, indicating that additional molecular diversity enhances representation learning.  
Second, GeoRecon remains effective even in low-resource regimes; when pretrained with only $10$k samples, the performance degradation is moderate, and the model still outperforms several full-data baselines. Notably, the $10$-layer encoder with $\lambda=1.00$ on the $100$k subset achieves the best finetuning performance, surpassing MolSpectra and Coord models pretrained on the full PCQM4Mv2 dataset.  

These findings highlight that GeoRecon is robust to reduced dataset scale while benefiting from additional data, making it particularly attractive for scenarios where only limited pretraining resources are available. 
\label{app:de}

\section{Baselines}
\label{app:bsl}
\textbf{Training from scratch}: {SchNet}~\citep{schutt2018schnet} employs continuous-filter convolutions for quantum interactions. {EGNN}~\citep{Satorras2021} introduces equivariant message passing under Euclidean transformations. {DimeNet}~\citep{gasteiger2020directional} and {DimeNet++}~\citep{gasteiger2020fast} leverage angular information, with the latter improving efficiency and accuracy. {PaiNN}~\citep{schutt2021equivariant} extends equivariant message passing to tensorial targets. {SphereNet}~\citep{liu2021spherical} incorporates spherical coordinates, while {TorchMD-Net}~\citep{tholke2022torchmd} applies equivariant transformers to molecular dynamics.

\textbf{Pretraining then finetune:} {Transformer-M}~\citep{luo2022one} encodes both 2D and 3D molecular data. {SE(3)-DDM}~\citep{liu2022molecular} adopts SE(3)-invariant denoising distance matching. {3D-EMGP}~\citep{jiao2023energy} leverages energy-based equivariant objectives. {Coord}~\citep{Zaidi2023} applies coordinate denoising for geometry-aware learning. 


\section{Training Information}
\label{app:t}
\textbf{Devices.}\, Experiments on MD17 and QM9 were conducted using NVIDIA RTX 4090 GPUs. Pretraining was performed on 4 $\times$ RTX 4090 GPUs (24 GB each) for 400k steps, while downstream fine-tuning for each task was carried out on a single RTX 4090 GPU. For MD22, fine-tuning experiments were conducted on a single NVIDIA A100 GPU.

\textbf{Hyperparameter Settings.}\, Our hyperparameter settings are shown in Table~\ref{tab:training_config}, which closely follow those of Coord~\citep{Zaidi2023}, with minimal modifications that we reduce the pretraining batch size on the QM9 dataset from 70 to 50 due to GPU memory constraints.
\begin{table}[ht]
    \centering
    \caption{Training Configuration}
    \label{tab:training_config}
    \resizebox{0.7\textwidth}{!}{
    \begin{tabular}{l|cccc}
    \toprule
    \multirow{2}{*}{} & \multicolumn{2}{c|}{\textbf{Finetuning}} & \multicolumn{2}{c}{\textbf{Pretraining}} \\
    & QM9 & MD17 & QM9 & MD17 \\    \midrule
    batch\_size &            128 &               4 &        50 &        70 \\
    cutoff\_lower &              0 &               0 &         0 &         0 \\
    cutoff\_upper &              5 &               5 &         5 &         5 \\
    ema\_alpha\_dy &              1 &               1 &         1 &         1 \\
     ema\_alpha\_y &              1 &            0.05 &         1 &         1 \\
embedding\_dimension &            256 &             128 &       256 &       128 \\
   energy\_weight &              1 &             0.2 &         0 &         0 \\
    force\_weight &              1 &             0.8 &         1 &         1 \\
inference\_batch\_size &            128 &              64 &        70 &        70 \\
              lr &         0.0004 &          0.0005 &    0.0004 &    0.0004 \\
     lr\_schedule &  cosine\_warmup &  cosine\_warmup  &    cosine &    cosine \\
          lr\_min &      1e-7 &       1e-7 & 1e-7 & 1e-7 \\
     lr\_patience &             15 &              30 &        15 &        15 \\
 lr\_warmup\_steps &          10000 &            1000 &     10000 &     10000 \\
lr\_cosine\_length &         100000 &           20000 &    400000 &    400000 \\
max\_num\_neighbors &             32 &              32 &        32 &        32 \\
           max\_z &            100 &             100 &       100 &       100 \\
       num\_heads &              8 &               8 &         8 &         8 \\
      num\_layers &              8 &               6 &         8 &         6 \\
       num\_nodes &              1 &               1 &         1 &         1 \\
         num\_rbf &             64 &              32 &        64 &        32 \\
     num\_workers &              6 &               6 &         6 &         6 \\
       precision &             32 &              32 &        32 &        32 \\
   save\_interval &             10 &              10 &         1 &         1 \\
   test\_interval &             10 &             100 &         1 &         1 \\
position\_noise\_scale &              0 &           0.005 &      0.04 &      0.04 \\
denoising\_weight &            0.1 &             0.1 &         1 &         1 \\
    \bottomrule
    \end{tabular}}
    \end{table}

\textbf{Computational Cost.}\, We evaluated three models—Coord, GeoRecon, and Frad—under the same experimental setup using the TorchMD-Net backbone on a single NVIDIA RTX 4090 (24GB) GPU with a batch size of 25. The average per-step training times were 0.1057 s for Coord, 0.1867 s for GeoRecon, and 0.2200 s for Frad. Although GeoRecon incurs a higher training cost than Coord, it remains more efficient than Frad. The additional overhead primarily arises from encoding three molecular variants (clean, noised, and large-noise) at each training step, whereas Coord processes only the noised conformation. Importantly, this extra cost is incurred only during the one-time pretraining phase. Since the pretrained checkpoint can be reused across diverse downstream tasks, the practical impact of this overhead is limited, while the benefits in downstream performance are more consequential.

\textbf{Code Availability.}\, We have provided the codes in the supplementary materials.




\end{document}